\documentclass[12pt]{article}
\usepackage{mathptmx}
\usepackage[T1]{fontenc}
\usepackage{graphicx} 
\usepackage[style=apa, backend=biber, natbib]{biblatex}
\usepackage[margin=1in]{geometry}
\usepackage{algorithm}
\usepackage{algpseudocode}
\usepackage{multirow}
\usepackage{array}
\usepackage{booktabs}
\usepackage{adjustbox}
\usepackage{xcolor}
\usepackage{amsmath}
\usepackage{color}

\title{Dynamic Tiling: A Model-Agnostic, Adaptive, Scalable, and Inference-Data-Centric Approach for Efficient and Accurate Small Object Detection}
\author{Son The Nguyen$^1$, Theja Tulabandhula$^1$, Duy Nguyen$^2$}

\date{%
    $^1$University of Illinois at Chicago,
    $^2$AI Affinity LLC\\[2ex]%
}

\begin{document}

\maketitle
\begin{abstract}
We introduce Dynamic Tiling, a model-agnostic, adaptive, and scalable approach for small object detection, anchored in our inference-data-centric philosophy. Dynamic Tiling starts with non-overlapping tiles for initial detections and utilizes dynamic overlapping rates along with a tile minimizer. This dual approach effectively resolves fragmented objects, improves detection accuracy, and minimizes computational overhead by reducing the number of forward passes through the object detection model. Adaptable to a variety of operational environments, our method negates the need for laborious recalibration. Additionally, our large-small filtering mechanism boosts the detection quality across a range of object sizes. Overall, Dynamic Tiling outperforms existing model-agnostic uniform cropping methods, setting new benchmarks for efficiency and accuracy.
\end{abstract}

\section{Introduction}

Off-the-shelf deep learning-based object detection models often struggle to detect extremely small objects that are significantly smaller than the overall image size. To tackle this problem, previous studies have proposed various solutions such as modifying network architectures with scale-aware or attention mechanisms, artificially increasing image resolution using super-resolution techniques, or performing density cropping using additional annotations or auxiliary learning architectures \citep{9143165}. However, these methods can be resource-intensive, time-consuming, and practically complex; which may hamper the broader adoption of this technology. 

We argue that small object detection, particularly the handling of prediction data during inference, embodies a unique inference-data-centric challenge. This perspective diverges from the more generalized view of data-centric AI, which predominantly emphasizes refining and curating training datasets while maintaining a consistent approach in model architectures, hyperparameters, and training pipelines. However, we posit that a holistic data-centric approach should encompass the entire lifecycle of a model. Beyond just training, passive or imprecise handling of data during both development and, crucially, during the inference stage can introduce systematic data errors or inefficiencies. Simply put, an overemphasis on training data quality, without a corresponding focus on data handling during model development and inference, may undermine the effectiveness of the entire system. Consequently, a comprehensive approach that prioritizes accurate data representation throughout both development and inference is essential to realize truly impactful object detection models.

Tiling, or uniform cropping, emerges as a promising inference-data-centric approach for small object detection. During inference, this technique uniformly breaks down images into smaller patches, each subjected to independent object detection. These separate detections are later combined using Non-Maximum Suppression (NMS) to present a holistic detection collection of the entire image. Fixed overlaps are commonly utilized during tiling to ensure that objects on the borderlines of these patches are effectively captured by adjacent tiles. Full image inference can also be employed with the smaller patches using NMS to detect larger objects. Tiling increases the relative pixel coverage of smaller objects compared to simply using the original image as input to the model. It allows models to fully harness the source image's resolution, despite the constraints of limited GPU memory. Although this strategy has been critiqued for inefficiency due to performing detection on overlapping and empty patches, it is more straightforward compared to many alternatives and has seen widespread adoption because it can be used only in the inference phase. The only adjustment made during training to accommodate tiling better is in the data augmentation process. Consequently, object detectors that utilize tiling can demonstrate excellent performance in the immediate detection of small objects with minimal modifications.

While current methodologies lay a foundational approach, there exists an avenue for optimization to heighten their practicality. The limitations of a one-size-fits-all strategy become evident when uniformly applying an overlapping rate across all patches, as it inadequately caters to the diverse characteristics of objects of different sizes and different environments. Additionally, relying solely on NMS for blindly merging detections from different tiles often falls short, especially when faced with edge cases of irregularly shaped objects. Challenges are further compounded when large objects span multiple patches, resulting in fragmentation and subsequent detection inaccuracies. This becomes especially problematic for classes with closely resembling features, such as cars and trucks. The existing non-class-agnostic NMS, even when applied to both patches and full image predictions, struggles to address this. 

In response to these challenges, we introduce Dynamic Tiling, a flexible and adaptive tiling solution specifically tailored for small object detection during inference. Our method is built on the notion of dynamic overlapping rates, which are informed by initial detections from non-overlapping patches. This unique approach not only rectifies fragmented objects but also optimizes computational efficiency by selectively applying overlapping only when necessary.

\textbf{The key contributions of our work can be summarized as follows:}
\begin{itemize}
    \item \textbf{Dynamic Overlapping Rates:} By leveraging detections from initial non-overlapping patches, our dynamic overlapping rates offer a more effective way to resolve fragmented objects. This results in increased detection accuracy across a range of conditions.
    \item \textbf{Reduced Computational Overhead:} Dynamic Tiling reduces the number of forward passes through the object detection model by aggregating overlapping patches, leading to faster inference times without compromising detection quality.
    \item \textbf{Adaptability and Scalability:} Our method easily adapts to various operational environments, eliminating the need for laborious recalibration and ensuring seamless integration into existing pipelines.
    \item \textbf{Large-Small Filtering Mechanism:} Our large-small filtering mechanism consolidates detections from multiple patches as well as the entire image, enhancing the quality of object detection for both small and large objects.
\end{itemize}

\section{Related Work}
\noindent\textit{\textbf{Learning-based object detection:}}
In recent years, learning-based object detection techniques have gained widespread use across various applications. These techniques can be broadly classified into two primary categories: two-stage detectors and single-stage detectors. Two-stage detectors, such as Fast R-CNN \citep{7410526}, Faster R-CNN \citep{NIPS2015_14bfa6bb}, and Cascade R-CNN \citep{8578742}, involve an initial proposal stage that is subsequently refined. This process makes them slower but often yields more accurate results. In contrast, single-stage detectors, such as SSD \citep{10.1007/978-3-319-46448-0_2}, YOLO \citep{Jocher_YOLO_by_Ultralytics_2023}, and RetinaNet \citep{8237586}, bypass the proposal stage to directly predict object locations, resulting in faster, albeit less accurate, detection.

Recently, anchor-free detectors like FCOS \citep{9010746}, VFNet \citep{Zhang2020VarifocalNetAI}, TOOD \citep{9710724}, and the more recent Yolov8 \citep{Jocher_YOLO_by_Ultralytics_2023} have begun to garner attention. These models remove the need for predefined anchor boxes and strive to strike a balance between detection speed and accuracy. We undertake an empirical study using the anchor-free one-stage YOLOv8m detectors.

\noindent\textit{\textbf{Tiling-based solutions to small object detection:}} To address the challenge of small object detection, some prior studies have manipulated data during inference by dividing images into smaller patches, a process known as tiling.

For instance, DarkHelp for YOLOv4 \citep{Darkhelp} allows users to define a fixed tiling dimension (e.g., 3x2, 5x3). If detected objects are located near the "dividing line" (the boundary between two tiles) and the bounding boxes are approximately of similar size, these are likely parts of the same object split across two tiles. Consequently, they are "merged" into a single entity. However, this method can lead to incorrect detections at the boundaries. For example, two separate objects located adjacently and sharing similar sizes could be incorrectly merged. Similarly, if a single irregular object is divided into large and small boxes, these might not be recognized as belonging to the same entity.

The predefined dimension with overlapping tiling method \citep{tiling} requires users to define a fixed tiling dimension and an overlap threshold (e.g., 25\%, 50\%). However, using a fixed overlap with a specific number of tiles necessitates an increase in tile size, which could potentially decrease pixel coverage for small objects.

SAHI \citep{SAHI}, a seminal paper in the tiling approach, employs a user-specified patch size (e.g., \(640 \times 640\)) and an overlap threshold that is applied across the entire image. One challenge with this approach is the potential need for adjustments or the addition of extra patches to cover areas that remain when the image dimensions and stride do not align perfectly. Moreover, although SAHI can be applied solely during inference, the original paper utilizes up-scaling for both training and inference. This design choice may not fully reveal the actual performance capabilities of SAHI. In contrast, our method maintains the resolution at the tile size for both training and inference, enabling us to isolate the effects of our Dynamic Tiling approach. Remarkably, even under these constraints, our Dynamic Tiling approach still achieves performance metrics that surpass those reported for SAHI in their original publication.

Moreover, adapting current tiling methods like DarkHelp, the predefined dimension with overlapping tiling, or SAHI to new environments presents yet another challenge: the necessity for recalibration of overlap thresholds. These existing methods rely on user-specified or predetermined tiling dimensions and overlap percentages, which may not translate well when applied to different data distributions or object scales. The need to fine-tune these parameters for each new setting significantly hampers the scalability and generalizability of such approaches. For instance, what works for urban scenes with large, well-defined objects may fall short when applied to natural environments where objects can be smaller and more irregularly shaped. In contrast, our Dynamic Tiling approach is designed to adapt seamlessly across various conditions, reducing the need for laborious and time-consuming recalibration, and thereby offering a more robust and versatile solution.

Building on these points, our Dynamic Tiling pipeline, rooted in our inference-data-centric philosophy, serves as a paradigm shift in object detection methodologies.  Unlike traditional methods that lock users into labor-intensive recalibration cycles when adapting to new settings, our system’s adaptive nature streamlines this process. By performing inference on non-overlapping tiles and generating and merging dynamic patches only when necessary, we not only minimize computational overhead, heighten detection accuracy, but also ensure accurate object detection across a myriad of operational environments. This ability to self-adjust according to the data at hand also makes our approach uniquely scalable and generalizable, setting a new standard in the field of small object detection. Additionally, the incorporation of an object size filter enhances the quality of object detection for both small and large objects, offering a comprehensive solution that meets a broad range of detection needs.

\section{Proposed approach}

\begin{figure}[h]
\centering
\includegraphics[scale=1]{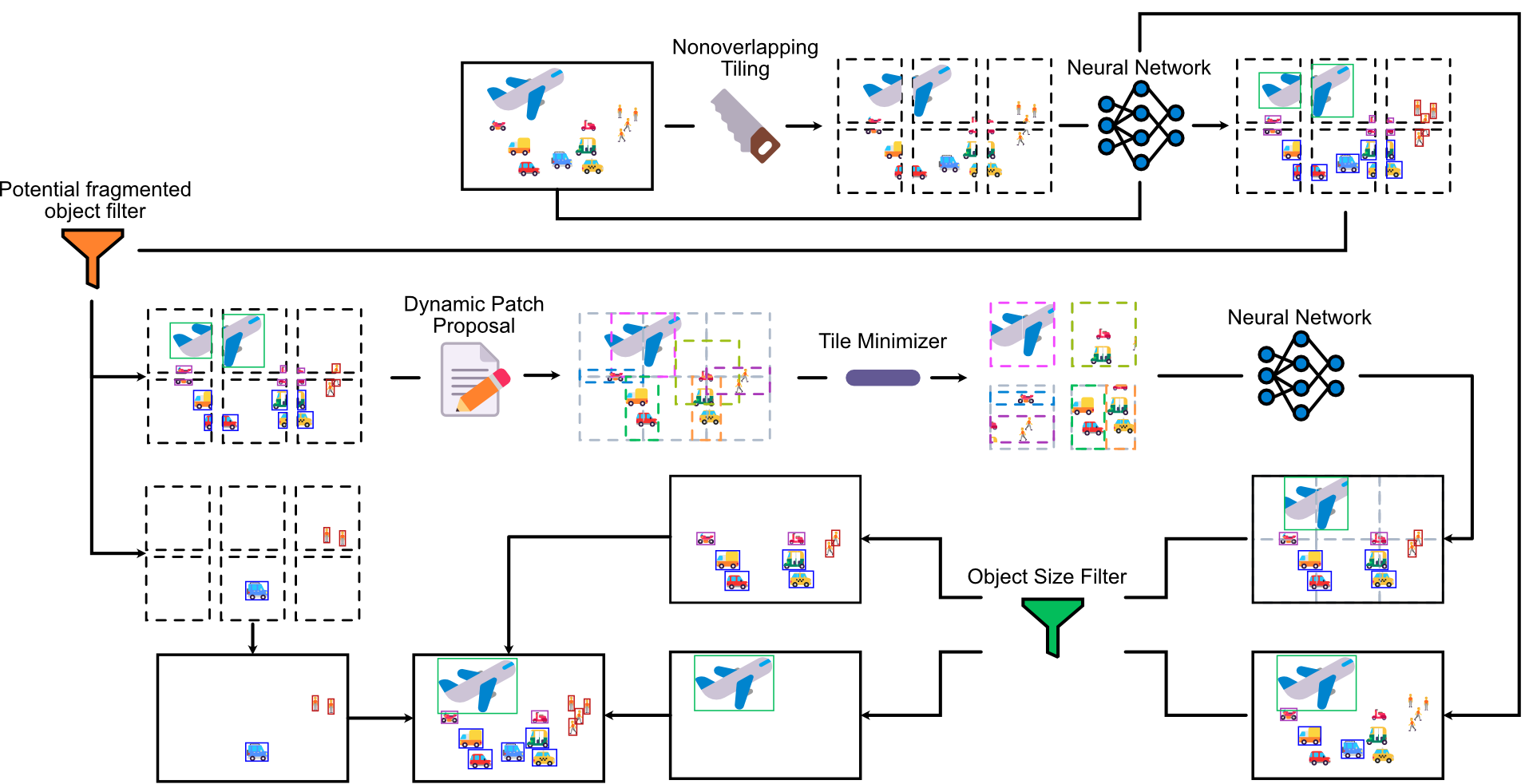}
\caption{Dynamic Tiling Procedure.}
\label{fig:dynamictilingdesign}
\end{figure}

We propose a novel Dynamic Tiling strategy, as illustrated in Figure \ref{fig:dynamictilingdesign}, that is centered around two key components. First, we uniformly divide an image into a predefined number of non-overlapping tiles and perform inference on them. Then, we used the predictions as guidance to generate supplementary dynamic patches on-demand to correct any potentially fragmented bounding boxes.

We start with an input image of dimensions \(W \times H\) and divide it into \(N \times M\) non-overlapping patches. The dimensions of each patch are \( \frac{W}{N} \times \frac{H}{M} \). These patches are processed in a single batch through our object detection model to generate initial bounding box predictions efficiently.


After that, we classify each bounding box prediction based on its proximity to the boundaries of its respective patch. The categories are:

\begin{itemize}
    \item \textbf{Near-Edge} (\textit{near\_edge}): Located close to an edge but not a corner.
    \item \textbf{Near-Corner} (\textit{near\_corner}): Adjacent to a corner.
    \item \textbf{Central} (\textit{central}): Positioned away from both edges and corners.
\end{itemize}

We introduce a threshold distance metric, calculated as \(x\% \times (W \text{ or} H)\), where \(x\%\) is a small percentage, which we set to 2\%. Predictions within this threshold of a single edge of a tile are classified as \textbf{Near-Edge}, while those within this threshold of two edges (corner) are classified as \textbf{Near-Corner}. Both \textbf{Near-Edge} and \textbf{Near-Corner} boxes are considered likely to be fragmented and are collectively represented as \( \text{Pred}(i, j) \) for adjacent patches \(P_i\) and \(P_j\) that share a boundary \(E(i, j)\). Predictions that do not meet these criteria are categorized as \textbf{Central} and are directly included in our final prediction set.


We then utilize \textbf{Near-Edge} or \textbf{Near-Corner} predictions to recover the full bounding box predictions by creating dynamic tiles or we called it Dynamic Patch Proposal. Based on the nature of the shared boundary \(E(i, j)\), we formulate dynamic tiles \(D(i, j)\):

\begin{itemize}
    \item \textbf{Common Edge}: \(D(i, j)\) features two parallel sides aligned with \(E(i, j)\), and its perpendicular sides are determined by \( \text{Pred}(i, j) \). The maximum size that \(D(i, j)\) can archive is \( \frac{W}{N} \times \frac{H}{M} \) with \(E(i, j)\) as its axis of symmetry. However, if \( \text{Pred}(i, j) \) appears only in \( P_i \) and not in \( P_j \), the tile is expanded to cover 50\% of the dimensions of \( P_j \).
    \item \textbf{Common Corner}: \(D(i, j)\) is designed to be symmetrical around \(E(i, j)\). The size of this tile is \( \frac{W}{N} \times \frac{H}{M} \), consistent with the dimensions of the original non-overlapping patches.
\end{itemize}

After constructing these dynamic tiles, they serve as new regions for the model to predict non-fragmented bounding boxes. These are also processed in batch for computational efficiency. \( \text{Pred}(i, j) \) predictions are discarded and replaced by the predictions of the dynamic tiles that are of that are near or crossing \(E(i, j)\).

Our Tile Minimizer strategy then aggregates these dynamically generated tiles into fewer composite images with the size of \( \frac{W}{N} \times \frac{H}{M} \). This process reduces the number of forward passes through the model, further enhancing computational efficiency.

Finally, we introduce a size-based filtering mechanism to enhance the precision of the object detection results. This strategy is especially beneficial when the image contains larger objects that share similar features, such as cars and vans. If individual tiles capture only small portions of these objects, the model may find it challenging to distinguish between them, leading to potential misclassification. This issue could, in turn, undermine the effectiveness of post-processing techniques like NMS. To address this, we define a size threshold \( \theta \), calculated as \( \alpha \times (W \times H) \), where \( \alpha \) is a predefined constant between 0 and 1. For downscaled, full-sized images, we include only those predictions larger than \( \theta \) in the final output. Conversely, for dynamic tiles, only predictions smaller than \( \theta \) are deemed valid. Our underlying hypothesis for this approach is that larger objects are more readily identified when viewed within a broader context, whereas smaller objects benefit from a more localized perspective for accurate detection.

\section{Experiments}
\subsection{Datasets}
Our method's efficacy is assessed using the VisDrone 2019 dataset \citep{9021986}, a benchmark widely recognized in the field of small object detection. The dataset is an extensive collection of 10,209 images, divided into 6,471 images for training, 548 for validation, 1610 for public testing (providing ground truth), and 1580 for challenge testing (not providing ground truth). The dataset boasts a diverse set of scenarios, detailed with bounding box annotations and labels for 10 object categories: pedestrian, person, car, van, bus, truck, motor, bicycle, awning-tricycle, and tricycle. The imagery presents a mix of both urban and rural environments, captured under varying lighting and weather conditions, from multiple viewpoints, and showcasing an assortment of objects per image. This wide-ranging representation positions it as an optimal benchmark for testing the performance of small object detection methodologies.

\subsection{Training setup}
We utilize dual NVIDIA TITAN X GPUs, each equipped with 12 GB of memory. We finetune the pre-trained YOLOv8m model with 71 epochs, choosing not to freeze any layers for flexibility. All images in the dataset are resized to dimensions of $1920 \times 1080$. In the training set, we generate 6 tiles of size $640 \times 640$ from each full-size image. This is achieved by cropping the original image into $3 \times 2$ non-overlapping patches. We also resize the original image to $640 \times 640$ and put it together with the non-overlapping patches for training. Mosaic augmentation is consistently applied throughout our training process until the last 10 epochs. Additionally, we utilize MixUp augmentation with a probability of $20\%$ and scale-up/down with a scaling factor ranging from $-20\%$ to $20\%$, as methods to diversify and enhance our training samples. For optimization during the training process, we employ the SGD optimizer, starting with an initial learning rate of $0.01$, momentum of 0.937 and weight decay of 0.0005. 

\subsection{Results}
\begin{figure}[h]
\centering
\includegraphics[scale=1.3]{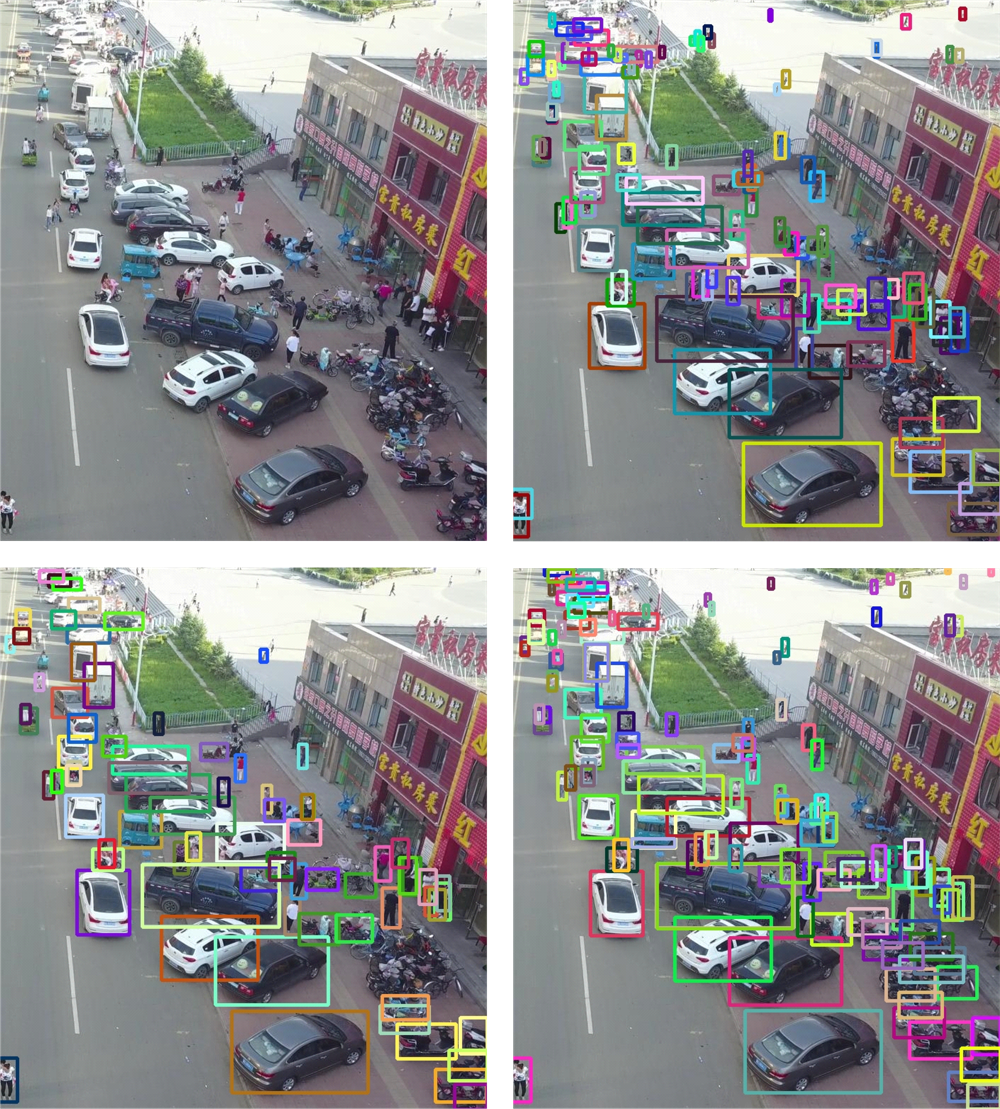}
\caption{An illustrative result of our adaptive tiling process. The top left shows the original image, while the top right presents the ground truth. The bottom left demonstrates the result of inference when only using the original image. The bottom right displays the result of inference when utilizing our Dynamic Tiling method.}
\label{fig:disparate-sized-objects-1}
\end{figure}

During inference, which is conducted on a single NVIDIA TITAN X GPU, we employ a $3\times2$ non-overlapping tiling grid as our initial tile size. Our method demonstrates high accuracy and efficiency in detecting small objects, substantiated by the illustrative results in figure \ref{fig:disparate-sized-objects-1}. Intriguingly, our approach proves so effective that it even identifies unlabeled objects—a common occurrence given the challenge of labeling small, densely packed items. This situation contributes to a decrease in mean average precision (mAP), affecting both our method and previous works.

Table \ref{tab:abc} provides a quantitative comparison of our method against earlier model-agnostic uniform cropping methods, presenting average inference times and standard deviations, mAP@50, mAP@50-95, mAP@50s (small-size box), mAP@50m (medium-size box), and mAP@50l (large-size box) following the Coco evaluation protocol \citep{coco}. The maximum number of detections allowed per image is 1000. We measure the efficiency of our approach by determining the average time it takes to predict one image. To provide a clearer understanding of these metrics, we present the details used for their calculations.
\begin{itemize}
    \item \textbf{Average Inference Time:} Calculated as the total time taken for inference divided by the number of images:
    \[
    t_{\text{avg}} = \frac{1}{N} \sum_{i=1}^{N} t_i
    \]
    where \( t_i \) is the time taken for the \(i^{\text{th}}\) image, and \(N\) is the total number of images.
    \item \textbf{Standard Deviation of Inference Time:} Calculated using the formula,
    \[
    \sigma = \sqrt{\frac{1}{N} \sum_{i=1}^{N} (t_i - t_{\text{avg}})^2}
    \]
    \item \textbf{mAP@50:} Mean Average Precision calculated at an Intersection over Union (IoU) threshold of 0.5 serves as a robust metric for object detection efficacy.
    \[
    \text{mAP@50} = \frac{1}{Q} \sum_{q=1}^{Q} \frac{\text{TP}_q}{\text{TP}_q + \text{FP}_q} \quad (\text{subject~to~IoU} \geq 0.5)
    \]
    where \( \text{TP}_q \) is the number of True Positives, \( \text{FP}_q \) is the number of False Positives, and \( Q \) is the total number of queries, for detections with IoU \( \geq 0.5 \) with ground truth.
    \item \textbf{mAP@50:95:} This is the average mAP calculated at different IoU thresholds from 0.5 to 0.95 with a step size of 0.05.
    \item \textbf{mAP@50s, mAP@50m, and mAP@50l:} These are the mAP calculated at an IoU of 0.5 but specifically for objects with areas less than \(32^2\) for small, between \(32^2\) and \(96^2\) for medium, and greater than \(96^2\) for large respectively.
\end{itemize}

\begin{table}[!htbp]
\centering
\caption{Comparison of proposed and competing methods on the VisDrone 2019 test set \citep{9021986}. FI stands for Full-image Inference. TTA stands for Test-time Augmentation.}
\adjustbox{max width=\textwidth}{
\begin{tabular}{l *{8}{c}}
\toprule
Model type & \multicolumn{8}{c}{VisDrone 2019 test set}\\
\cmidrule(lr){2-9}
{} & Model size & Avg. Inference Time & STD & mAP@0.5 & mAP@0.5:0.95 & mAP@0.5s & mAP@0.5m & mAP@0.5l \\
\midrule
YOLOv8m + FI & 640 & 0.017 & 0.059 & 26.8 &  16.8 &  9.8 & 39.7 & 54.3 \\
YOLOv8m + FI + TTA & 640 & 0.031 & 0.006 & 27.8 & 17.2 & 10.3 & 40.8 & 58.3  \\
YOLOv8m + Darkhelp & 640 & 0.22 & 0.373 & 39.5 & 23.1 & 26 & 50.9 & 49.8 \\
YOLOv8m + Darkhelp + TTA & 640 & 0.313 & 0.427 & 41.6 & 24.3 & 27.5 & 53.3 & 57.5 \\
YOLOv8m + PowerTiling + FI & 640 & 0.135 & 0.036 & 39.6 & 23.8 & 24.3 & 52.4 & 58.3 \\
YOLOv8m + PowerTiling + FI + TTA & 640 & 0.24 & 0.04 & 40.4 & 24.2 & 24.9 & 53.5 & 61 \\
YOLOv8m + SAHI & 640 & 0.264 & 0.011 & 39.5 & 24 & 25.4 & 51.1 & 52.1 \\
YOLOv8m + SAHI + TTA & 640 & 0.447 & 0.018 & 41.8 & 24.8 & 26.8 & 53.8 & 59.4 \\
YOLOv8m + SAHI + FI & 640 & 0.305 & 0.015 & 40.2 & 23.9 & 25.6 & 52.1 & 58.6\\
YOLOv8m + SAHI + FI + TTA & 640 & 0.503 & 0.022 & 41.6 & 24.7 & 26.8 & 53.9 & 61.8 \\
YOLOv8m + Dynamic Tiling (Ours) & 640 & 0.126 & 0.029 & 41.4 & 24.7 & 27.7 & 53.2 & 53.9 \\
YOLOv8m + Dynamic Tiling (Ours) + TTA & 640 & 0.16 & 0.040 & 43.5 & 26.0 & 28.6 & 55.7 & 61.3 \\
YOLOv8m + Dynamic Tiling (Ours) + FI & 640 & 0.160 & 0.032 & 43.8 & 26.1 & 29.9 & 55.6 & 55.1 \\
YOLOv8m + Dynamic Tiling (Ours) + FI + TTA & 640 & 0.333 & 0.061 & 45.6 & 27.1 & 31.3 & 57.7 & 60.7 \\
YOLOv8m + Dynamic Tiling (Ours) + FI + Tile Minimizer & 640 & \textbf{\textcolor{blue}{0.148}} & 0.025 & \textbf{\textcolor{blue}{43.8}} & \textbf{\textcolor{blue}{26}} & \textbf{\textcolor{blue}{29.9}} & \textbf{\textcolor{blue}{55.4}} & \textbf{\textcolor{blue}{54.9}} \\
YOLOv8m + Dynamic Tiling (Ours) + FI + Tile Minimizer + TTA & 640 & \textbf{\textcolor{blue}{0.301}} & 0.047 & \textbf{\textcolor{blue}{45.5}} & \textbf{\textcolor{blue}{27}} & \textbf{\textcolor{blue}{31.2}} & \textbf{\textcolor{blue}{57.5}} & \textbf{\textcolor{blue}{60.3}} \\
\bottomrule
\end{tabular}
}
\label{tab:abc}
\end{table}

Our approach, exemplified by our best configuration "YOLOv8m + Dynamic Tiling (Ours) + FI + Tile Minimizer + TTA," has set new standards in the domain of model-agnostic small object detection, as evidenced by its outstanding performance across various evaluation metrics. Specifically, our best configuration achieves a mAP@0.5 score of 45.5 and a mAP@0.5:0.95 score of 27. The inclusion of Full-Image Inference (FI) allows our model to take into account global contextual information when detecting objects, thereby improving accuracy. Its incorporation into our best configuration contributes significantly to the elevated mAP scores, especially when detecting both large and small objects in complex scenes. When compared to "YOLOv8m + SAHI + FI + TTA," which had a mAP@0.5 of 41.6 and a mAP@0.5:0.95 of 24.7, our best configuration shows a relative improvement of 9.4\% and 9.3\% in these metrics, respectively. This level of performance suggests that our approach, instantiated by Dynamic Tiling, offers a superior balance of precision and recall across a range of object sizes and IoU thresholds. Furthermore, the significant gains in our best configuration can be primarily attributed to its proficiency in detecting small objects, where it achieves a mAP@0.5s of 31.2. In comparison to other methods such as "YOLOv8m + SAHI + FI + TTA," "YOLOv8m + PowerTiling + FI + TTA," and "YOLOv8m + Darkhelp + TTA," our best configuration shows a relative improvement in mAP@0.5s of 16.4\%, 25.3\%, and 13.5\%, respectively.

It's important to note that there appears to be a trade-off between inference time and performance in our best configuration. While TTA enhances the mAP@0.5 score from 43.8 to 45.6, it more than doubles the average inference time from 0.160 to 0.333 seconds. However, Tile Minimizer, a feature in our best configuration, allows us to maintain high accuracy—comparable to configurations without Tile Minimizer—while notably reducing inference times. Specifically, the average inference time is brought down to 0.148 seconds, representing an improvement of 7.5\% over the 0.160-second TTA-exclusive configuration, and to 0.301 seconds, representing an improvement of 9.6\% over the 0.333-second TTA-inclusive configuration. Moreover, when compared to other previous methods, our best configuration consistently outperforms them in terms of inference time. Together with its outperformance in mAP metrics, this offers a more balanced solution for practical applications. Therefore, both our approach and our best configuration distinguish themselves by achieving remarkable accuracy while maintaining computational efficiency.

These collective findings unequivocally validate the efficacy of both our approach and our best configuration in establishing new benchmarks in model-agnostic small object detection in terms of computational efficiency and overall accuracy.

\section{Conclusion}
Existing tiling methods confront critical limitations such as difficulties in detecting objects of varying sizes and extended inference times. To address these challenges, we introduce Dynamic Tiling. This innovative method, rooted in our inference-data-centric paradigm, diverges from traditional techniques that mainly focus on neural network architectures and optimization strategies.

Our premier configuration, "YOLOv8m + Dynamic Tiling (Ours) + FI + Tile Minimizer + TTA," sets new benchmarks across multiple evaluation metrics, including mAP@0.5, mAP@0.5:0.95, and particularly mAP@0.5s for small object detection. Notably, our approach significantly increases detection accuracy through the use of dynamic overlapping rates, which enable the effective rectification of fragmented objects. In addition to improved accuracy, Dynamic Tiling reduces inference time by minimizing the number of forward passes needed through the object detection model and avoiding unnecessary overlapping in patches. This reduction in computational overhead strikes a unique balance between efficiency and performance, making our approach particularly suited for real-time applications.

What sets our method apart is its unparalleled adaptability. Dynamic Tiling eliminates the need for laborious recalibration, offering seamless integration into existing pipelines across a wide range of operational environments. This exceptional adaptability is further augmented by a large-small filtering mechanism that effectively consolidates detections from both patches and full images, enhancing the detection quality for objects of all sizes.

Given these advantages, Dynamic Tiling does not merely find immediate applicability but also serves as a catalyst for revolutionary advancements in the fields of real-time small object detection and spatial computing, both of which require a delicate balance of computational power and accuracy. Through our focus on inference-data-centricity, we anticipate that Dynamic Tiling will establish a new paradigm, acting as a cornerstone for future innovations and setting unprecedented standards in object detection.

\clearpage
\printbibliography

\end{document}